\documentclass[10pt,twocolumn,letterpaper]{article}

\usepackage[draft,frozencache=true,cachedir=tmp]{minted}
\usepackage{listings}

\usepackage{iccv}
\usepackage{times}
\usepackage{epsfig}
\usepackage{graphicx}
\usepackage{amsmath}
\usepackage{amssymb}
\usepackage[numbers,sort,compress]{natbib}
\usepackage{booktabs}
\usepackage[pagebackref=true,breaklinks=true,letterpaper=true,colorlinks,bookmarks=false]{hyperref}

\usepackage[pagebackref=true,breaklinks=true,letterpaper=true,colorlinks,bookmarks=false]{hyperref}

\iccvfinalcopy 

\def\iccvPaperID{5} 

\ificcvfinal\fi

\begin{document}

\title{SSR: Semi-supervised Soft Rasterizer for single-view 2D to 3D Reconstruction}

\author{Issam Laradji\\ 
McGill University, Element AI\\
{\tt\small issam.laradji@gmail.com} \and
Pau Rodríguez \\
Element AI \\
{\tt\small pau.rodriguez@servicenow.com} \and
David Vazquez \\
Element AI \\
{\tt\small david.vazquez@servicenow.com} \and
Derek Nowrouzezahrai\\
McGill University\\
{\tt\small derek@cim.mcgill.ca}

}

\maketitle
\ificcvfinal\thispagestyle{empty}\fi

\begin{abstract}
Recent work has made significant progress in learning object meshes with weak supervision. Soft Rasterization methods have achieved accurate 3D reconstruction from 2D images with viewpoint supervision only. In this work, we further reduce the labeling effort by allowing such 3D reconstruction methods leverage unlabeled images. In order to obtain the viewpoints for these unlabeled images, we propose to use a Siamese network that takes two images as input and outputs whether they correspond to the same viewpoint. During training, we minimize the cross entropy loss to maximize the probability of predicting whether a pair of images belong to the same viewpoint or not. To get the viewpoint of a new image, we compare it against different viewpoints obtained from the training samples and select the viewpoint with the highest matching probability.  We finally label the unlabeled images with the most confident predicted viewpoint and train a deep network that has a differentiable rasterization layer. Our experiments show that even labeling only two objects yields significant improvement in IoU for ShapeNet when leveraging unlabeled examples. Code is available at \url{https://github.com/IssamLaradji/SSR}.
\end{abstract}

\begin{figure*}[t]
\begin{center}
   \includegraphics[width=0.8\linewidth]{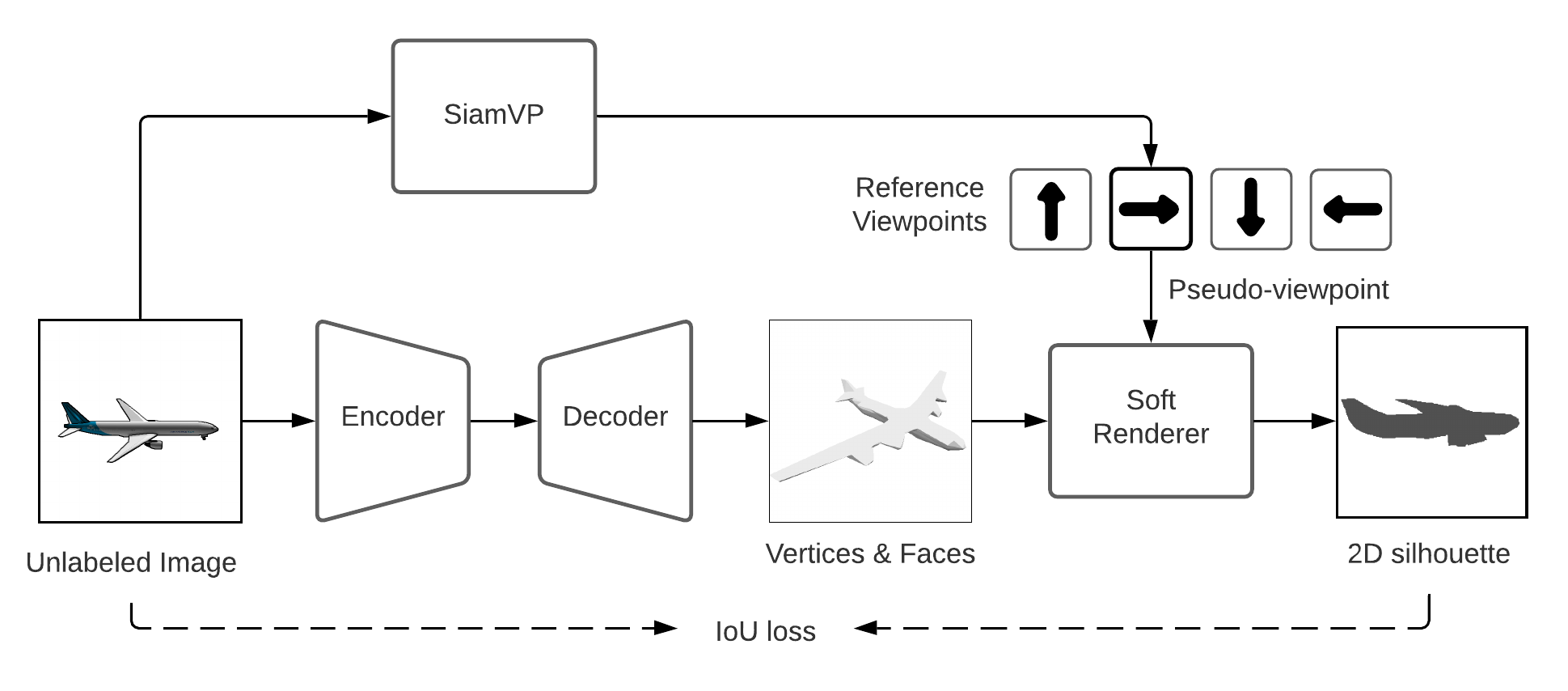}
\end{center}
   \caption{SSR consists of two main components: SiamVP and SoftRas (Soft Renderer in the picture). The SiamVP learns whether two images have the same viewpoint whereas SoftRas learns to project the input image into a silhouette based on a selected viewpoint.}
\label{fig:softras}
\end{figure*}

\section{Introduction} 
Acquiring 3D representations of objects is key in many  application domains. For example, these representations can be used in  autonomous driving, robotics, remote sensing, and medical treatment~\citep{chen2017multi,guo2020deep}. They can also be used to create digital twins of dynamic scenes for deep analysis~\citep{el2018digital}, and to synthesize novel views and scenes as efficient ways to augment datasets~\citep{liu2018semi}. Manually modelling 3D objects requires a significant amount of human effort. Fortunately, we can reduce that effort using models that combine deep learning and differentiable renderers to train models on 2D images of an object and generate a 3D representation of that object~\cite{liu2019soft}.

Over the last years, the research community has proposed multiple methods to learn to perform single-view 2D image to 3D reconstruction using either viewpoints, voxels, point-clouds, or silhouttes~\citep{cashman2012shape, choy20163d, delanoy20183d, ulusoy2015towards}. However, they assume that the dataset is fully labeled and they do not leverage unlabeled examples which are available in abundance. Such additional data could potentially provide strong signal for the model to boost its 3D reconstruction performance. Although work in this area is still scarce~\citep{rezende2016unsupervised, liao20183d, Rajeswar2020Pix2ShapeTU}, in other areas like classification and semantic segmentation using unlabeled data is a common practice and is proven to work with techniques based on semi- and self-supervision~\cite{sohn2020fixmatch, xie2020self, chen2020improved, manas2021seasonal, chen2020simple, grill2020bootstrap, Rodriguez2020EmbeddingPS}.

\begin{figure}[!t]
\begin{center}
   \includegraphics[width=0.8\linewidth]{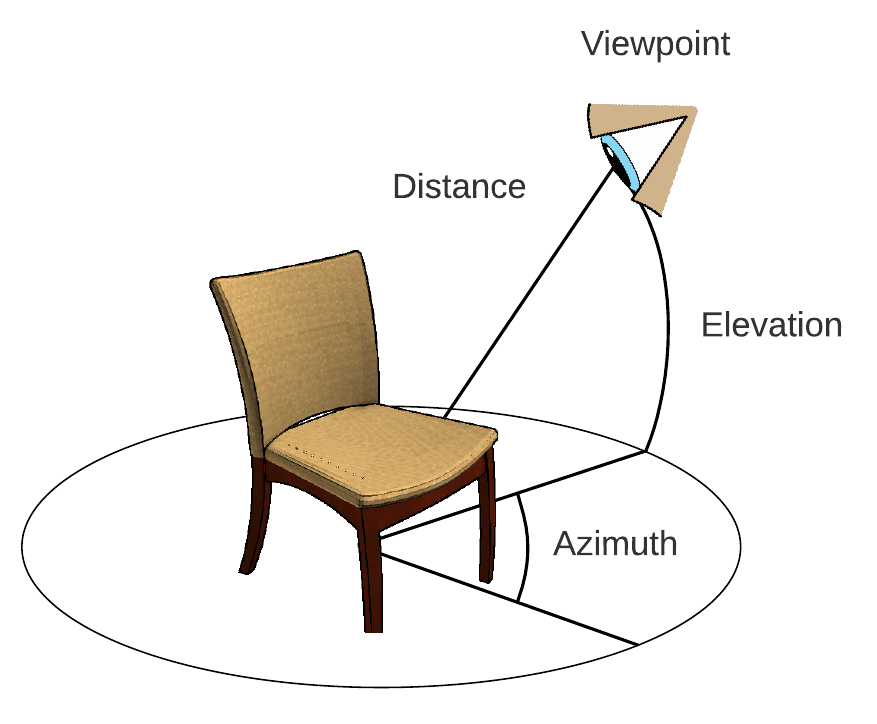}
\end{center}
   \caption{Illustration of the viewpoint coordinate system. The azimuth is the horizontal angle between the camera and the  of the object. Elevation is the angle with the horizontal plane. Distance is the euclidean distance between the camera and the the object.}
\label{fig:viewpoint}
\end{figure}

Semi-supervised methods for single-view 2D image to 3D reconstruction with viewpoint supervision remain largely unexplored. A viewpoint represents the distance, elevation, and azimuth from a set vantage point as shown in Figure~\ref{fig:viewpoint}. The most relevant work is~\citep{liao20183d} which addresses this task with 3D supervision. Unfortunately, 3D supervision is costly to acquire and it often requires 3D modeling expertise for creating the CAD models of different objects, which could be prohibitive in many cases. Viewpoint supervision on the other hand is easier to acquire. For instance, it is possible to obtain viewpoints from well-calibrated cameras~\citep{arkit,nyimbili2016structure}. However, requiring calibration makes it difficult to acquire labeled training data. In other cases, it is possible to train a generic viewpoint inference model~\citep{mustikovela2020self} that could be used out of the box to get viewpoint estimates from images in the wild. The advantage of these scenarios is that they do not need 3D expertise in the process of learning to construct 3D meshes. Although viewpoints are much easier to acquire than 3D models of an object, their collection is not straightforward for objects in specific domains.

In this paper, we address the problem of 3D reconstruction from a single 2D image without 3D supervision. We propose SSR, a semi-supervised learning algorithm to leverage unlabeled data and reduce the viewpoint annotation costs such as camera calibration or human expertise. In order to approximate viewpoints from unlabeled data we propose SiamVP, a Siamese network~\citep{chopra2005learning} that is trained to take two images with known viewpoint as input and outputs the probability that those images correspond to the same viewpoint. To approximate the viewpoints of unlabeled images, we use SiamVP to compare them against other training images of the same class with known viewpoints. We finally assign the most confident predicted viewpoints to the unlabeled images and train the soft rasterizer on them. In addition, we also include augmentation to our training of SiamVP. Two images of matching viewpoints will still match when they undergo the same affine transformation such as rotation, scale and translation. We empirically show that SSR achieves significant improvement in IoU for ShapeNet~\cite{chang2015shapenet} when leveraging unlabeled samples even when only two labeled objects are available. In an ablation study, we empirically show the effectiveness of the proposed data augmentation based on rotations for SiamVP.

The contributions of this work can be summarized as follows:
\begin{itemize}
    \item We establish a novel framework that leverages unlabeled samples for learning to construct 3D meshes from 2D images with viewpoint supervision.
    \item We show that using augmentations can highly stabilize the training of SiamVP and achieve more accurate viewpoint predictions.
    \item We show that SSR provides a consistent boost of performance over SoftRas for different amounts of labeled data with respect to IoU on ShapeNet.
\end{itemize}

\section{Related Work}
This work intersects with the topics of Reconstruction with 3D supervision, Reconstruction with Viewpoint Supervision, and Semi- and Self-Supervised Learning which we will discuss in detail in the following paragraphs.

\paragraph{Reconstruction with 3D supervision}
Many deep learning based 3D shape reconstruction methods in the literature require the 3D model ground truth to be observed during training~\citep{choy20163d, girdhar2016learning, wang2018pixel2mesh, gkioxari2019mesh}. 
\citet{girdhar2016learning} propose to learn a joint embedding for both 3D shapes and 2D images with full supervision. Likewise, methods such as Pixel2Mesh~\citep{wang2018pixel2mesh} and Mesh R-CNN~\citep{gkioxari2019mesh} are trained to reconstruct mesh vertices by minimizing the error with respect to the ground truth 3D model. Methods such as O-Net~\citep{mescheder2019occupancy} predict if randomly sampled 3D points are inside or outside the 3D models instead of directly approximating their vertices. These methods obtain high reconstruction accuracy. Unfortunately, the cost of acquiring 3D ground truth annotations is high, which motivates research for alternative approaches with weaker supervision such as 2D images and viewpoints.

\paragraph{Reconstruction with Viewpoint Supervision}
More recently, some works avoid 3D supervision by taking advantage of differentiable renderers~\citep{kato2018neural, liu2019soft, chen2019learning} with either multiple views, or known ground truth camera poses.
\citet{yan2016perspective} leverage perspective transformations and the ground truth viewpoints in order to recover the 3D objects corresponding to the 2D images without 3D ground truth information. Similarly, \citet{gwak2017weakly} leverage 2D perspective projections while constraining the reconstructed 3D shapes to a manifold of 3D unlabeled shapes of real-world objects. \citet{arsalan2017synthesizing} proposed to remove the viewpoint supervision by introducing depth maps and learning a generative model over 2D projections and depth maps, thus requiring both types of input to recover 3D models at test time. \citet{tulsiani2018multi} introduce a consistency loss that, given two images of an object instance with two different poses, enforces that the shape predicted for one pose is consistent with the depth/mask observation of the second image. Although these works reduce the need for 3D annotations, they still require full supervision with a weaker form of label such as viewpoints. In this work, we propose to automatically approximate viewpoints from unlabeled data thus reducing the amount of required annotations.

\paragraph{Semi- and Self-Supervised Learning}
Leveraging unlabeled data for 3D reconstruction is largely unexplored. \citet{liao20183d} proposed a semi-supervised consistency loss term between 2D shapes and 3D objects. However, they still require 3D supervision during training. Likewise, \citet{piao2019semi} leveraged generative adversarial networks (GAN)~\citep{goodfellow2020generative} for semi-supervised 3D facial modelling with limited 3D supervision. In order to remove 3D supervision, \citet{li2020self} proposed a self-supervised approach that leveraged semantic knowledge between object parts. Thus, their method cannot be applied on completely unlabeled data where part-annotations are not present like the problem we study in this work.

\section{Methodology}
\label{sec:methodology}
\begin{figure*}[!t]
\begin{center}
   \includegraphics[width=1.0\linewidth]{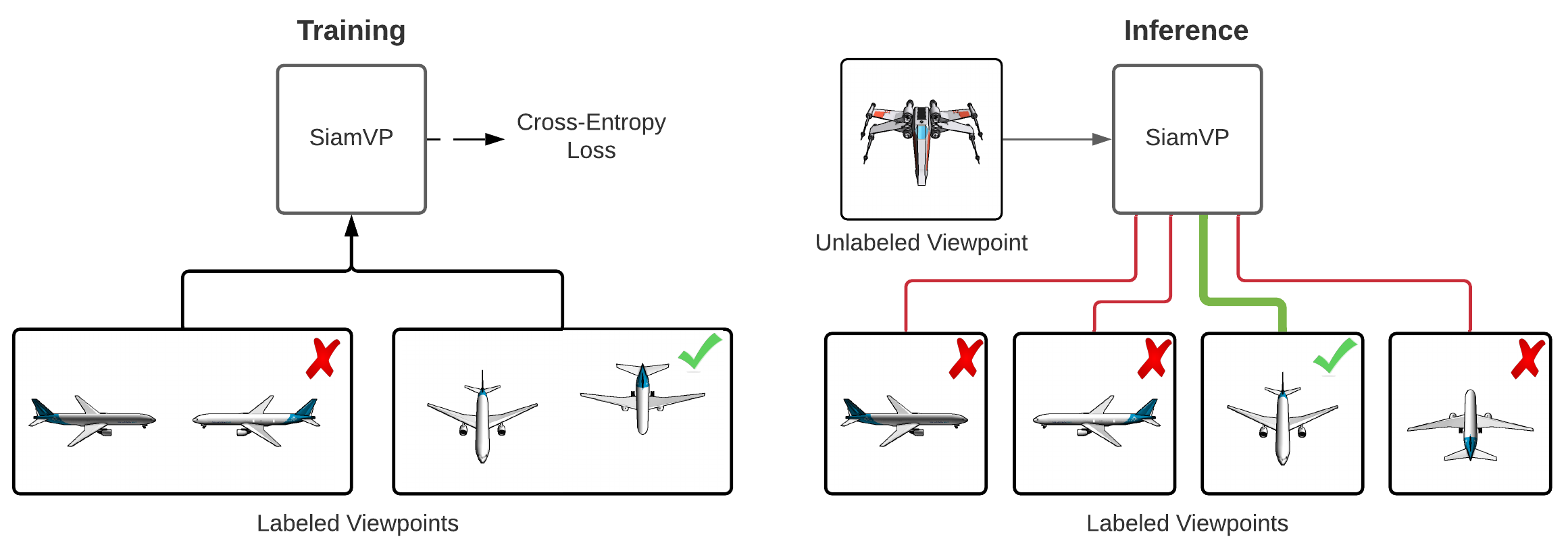}
\end{center}
   \caption{Illustration of SiamVP. During training (left) SiamVP is optimized to infer whether two images correspond to the same viewpoint. During semi-supervised training (right), SiamVP is used to infer the viewpoint of new images with unkown viewpoint by matching their viewpoint with other images of known viewpoint.}
\label{fig:siamvp}
\end{figure*}

In this paper we propose SSR, a framework for semi-supervised 3D object reconstruction from 2D images. In order to reduce the amount viewpoint supervision, we propose to train a Siamese network (SiamVP) to infer the viewpoint for images for which the viewpoint is unknown. SiamVP takes two images as input and identifies whether they are acquired from the same viewpoint (Figure~\ref{fig:siamvp}). This network is then used to predict viewpoint pseudo labels for the unlabeled set by comparing them with images with a known viewpoint. These images and the inferred viewpoints are used to train a soft rasterizer network (Figure~\ref{fig:softras}).  In the next sections we describe the training data, the pipeline that involves optimizing the soft rasterizer network, the Siamese viewpoint predictor and the training and validation procedure.

\subsection{Training Data}
We consider the standard semi-supervised learning setting~\citep{chapelle2009semi}, where there is a set of labeled images $X_L$ whose labels are the viewpoints and a set of unlabeled images $X_U$ with unknown viewpoint. We assume that the same objects can appear in multiple images under different viewpoints both in $X_L$ and $X_U$ and that we know the sets of images that correspond to each object. For each image we have the foreground mask for the object of interest, in some cases like in ShapeNet the input image is also considered the foreground mask. These assumptions are common for this problem setup~\cite{tulsiani2018multi,umr2020}.

\subsection{Training Pipeline}
As in Figure~\ref{fig:softras}, training consists of optimizing both SiamVP and SoftRas. We describe each of the two components in the following sections. 

\paragraph{SoftRas.} We train a model for 2D to 3D reconstruction using the procedure described by~\citet{liu2019soft}. First, a 2D image $I_s$ is provided as input to the SoftRas model. The model consists of an encoder and a decoder and outputs a 3D mesh $M$ composed of vertices and faces. That 3D representation is projected into a silhoutte $\hat{I}_s$ with the viewpoint associated with that image using a soft rasterizer. Our loss consists of the following two terms: $\mathcal{L}_s + \mathcal{L}_g$. The $\mathcal{L}_s$ term corresponds to the silhouette loss:
\begin{equation}
\label{eq:iou}
    \mathcal{L}_s = 1 - \frac{||\hat{I}_s * I_s||_1}{||\hat{I}_s + I_s - \hat{I}_s * I_s||_1},
\end{equation} where * and + are the
element-wise product and sum operators, respectively. $\mathcal{L}_g$ is the geometric loss that regularizes the Laplacian of the shape. Note that this model is amenable to any encoder-decoder architectures. Its key component is the soft rasterizer function that allows the gradients to flow from the IoU loss computed in Eq.~\ref{eq:iou} to the parameters of the encoder and decoder.

\paragraph{SiamVP.} At each iteration SiamVP takes pairs of images, where some of them  correspond to the same viewpoint and others do not (Figure~\ref{fig:siamvp}). Pairs of images of the same viewpoint are sampled from different objects to avoid the model optimizing a trivial loss. Since most pairs of images have different viewpoints, we sample a balanced set of pairs from images with same and different viewpoints and also mine hard negatives and positives to ensure a stable loss optimization. Hard negatives are those pairs that have dissimilar viewpoints and have the highest predicted probability of being the same, whereas hard positives are those pairs that have the same viewpoints but have the lowest predicted probability of being the same. This procedure is also shown in Step 4 of Algorithm 1.

Thus given two images $I_i$ and $I_j$ and an indicator function $s_{i,j}$ that is 1 when the viewpoint of $I_i$ is the same as the viewpoint of $I_j$ and $0$ otherwise, we minimize the following loss function:


\begin{equation}
\label{eq:siamvp}
    \mathcal{L}_{v_{(I_i, I_j)}} =  - s_{i,j}\log(P_v) - (1 - s_{i,j})\log(1 - P_v).
\end{equation}

In the case where only few examples have been labeled, it is important to augment the dataset to avoid overfitting. Between pairs of images with same viewpoint, we also augment the pair with identical random rotation between 0 and 360 degrees and incorporate their similarity probability into the loss in Eq.~\ref{eq:siamvp}.

 
\paragraph{Pseudo-Labeling.}
While SiamVP is being trained, we add the viewpoint labels to the unlabeled set of $N_u$ examples as in Step 6 of Algorithm~\ref{listing}. The "siam\_vp" predict function computes the viewpoint of an unlabeled image as follows. First, from the labeled set we randomly select an image from each viewpoint as reference and compare it with each unlabeled example using SiamVP to get a similarity matrix $S \in \mathcal{R}^{N_u x N_v}$ where $N_v$ is the number of unique viewpoints we have in the training set. For each row $S_i$ we select the viewpoint $v_j$ with the highest probability $S_{ij}$. We also compute $\hat{S}$ which is the similarity matrix between the labeled and unlabeled examples after they undergo the same rotation. Like with $S$, we also select the viewpoint $\hat{v}_j$ with the highest probability  $\hat{S}_{ij}$ for each row $i$. If $S_{ij}$ and $\hat{S}_{ij}$ are both higher than $0.5$ and $v_j=\hat{v}_j$ then we select $v_j$ to be the label of unlabeled image $I_i$ otherwise we skip labeling the image. In our experiments, we observe that this approach tends to label a small portion of the unlabeled set each time this procedure is executed, and the accuracy between these predicted viewpoints and the groundtruth is often high.
\begin{table*}[!t]
\caption{Number of objects for each of the 13 ShapeNet Categories.}
\label{tab:stats}
\footnotesize
\centering
\begin{tabular}{lllllllllllll}
\toprule
      Plane &  Bench & Cabinet & Car  &  Chair & Display 
     & Lamp & Speaker 
     & Rifle & Sofa & Table & Phone & Vessel \\
\midrule
\midrule
4045 & 1816 & 1572 & 7496 & 6778 & 1095 & 2318 & 1618 & 2372 & 3173 & 8509 & 1052 & 1939  \\
\bottomrule
\end{tabular}
\end{table*}

\paragraph{The Full Cycle.}
Every $z$ iterations of training SiamVP and Softras, we execute the pseudo-labeling procedure from the previous paragraph to get more unlabeled examples labeled with viewpoints. In the first cycle where none of the unlabeled examples have been pseudo labeled, we train SoftRas using its standard procedure. In the subsequent cycles, each batch that SoftRas receives consists of half labeled examples, and half pseudo-labeled examples which are trained collectively using the SoftRas standard loss term which is shown in Step 3 in Algorithm~\ref{listing}. 

\paragraph{Inference.} At test time, SiamVP is discarded and the trained SoftRas takes as an input a single unseen 2D image of an object and outputs the 3D mesh of that object (Figure~\ref{fig:siamvp}).

\begin{table*}[!t]
\caption{IoU results for various objects of ShapeNet, where $N$ is the number of labeled examples.}
\label{tab:results}
\footnotesize
\centering
\begin{tabular}{llcccccccccccccc}
\toprule
     Method & N  & Plane &  Bench & Cabinet & Car  &  Chair & Display 
     & Lamp & Speaker 
     & Rifle & Sofa & Table & Phone & Vessel & Mean \\
\midrule
\midrule
MVC~\cite{tulsiani2018multi} &   0 & 0.38&  -&-&     0.48 &                    0.35 & -& -& -& -& -& -& -& -& -\\
\midrule
SoftRas &    2 & 0.42 &       0.24 &         0.39 &      0.6 &       0.26 &         0.33 &      0.17 &             0.38 &       0.44 &      0.38 &       0.17 &           0.57 &            0.28 &  0.36  \\

SSR (ours) &  2 &         0.55 &       0.44 &         0.48 &     0.75 &       0.37 &          0.5 &      0.27 &             0.45 &       0.63 &      0.53 &       0.27 &           0.71 &            0.53 &     0.50   \\

\midrule
SoftRas &    5 &  0.43 &       0.27 &         0.42 &     0.64 &       0.29 &         0.34 &      0.19 &             0.45 &       0.43 &      0.47 &        0.2 &            0.60 &            0.42 &  0.40  \\
SSR (ours)  &  5 &       0.59 &       0.44 &          0.5 &     0.75 &       0.49 &         0.55 &      0.33 &             0.47 &       0.65 &      0.66 &       0.35 &           0.78 &            0.56 &    0.55   \\
\midrule
SoftRas &    20 &  0.47 &       0.34 &         0.53 &     0.68 &       0.33 &         0.44 &      0.26 &             0.51 &       0.52 &      0.55 &       0.31 &           0.67 &            0.47 &  0.47 \\
SSR (ours)  &  20 &  0.63 &       0.47 &         0.65 &     0.76 &       0.51 &         0.57 &      0.38 &             0.62 &       0.66 &      0.65 &       0.43 &           0.78 &            0.55 &    0.59    \\
\midrule
\midrule
SoftRas & All &   0.64 &       0.51  &                0.71 & 0.77 & 0.53&0.62 & 0.46 & 0.67 & 0.68 & 0.69 & 0.45 & 0.79 & 0.60 & 0.62 \\
\bottomrule
\end{tabular}
\end{table*}

\begin{listing*}
\label{listing}
\caption{PyTorch-style pseudocode for SSR}
\small
\begin{minted}[frame=single,framesep=10pt,linenos]{python}
# softras: A SoftRas model
# siam_vp: A SiamVP model
for i in range(max_iter):
    # Step 1: randomly sample labeled batch
    images_labeled, viewpoints_labeled = labeled_set.sample()
    
    # Step 2: randomly sample pseudo-labeled batch
    images_unlabeled, viewpoints_unlabeled_pseudo = unlabeled_set.sample()
    
    # Step 3: Compute softras loss from Eq. 1
    images = torch.cat([images_labeled, images_unlabeled], dim=0)
    viewpoints = torch.cat([viewpoints_labeled, viewpoints_unlabeled_pseudo], dim=0)
    loss = compute_softras_loss(softras, images, viewpoints)
    
    # Step 4: Compute siamvp loss from Eq. 2
    image_pairs, viewpoint_pairs = get_pairs(images_labeled, viewpoints_labeled)
    loss = compute_siamvp_loss(siam_vp, image_pairs, viewpoint_pairs)
    
    # Step 5: optimization step
    optimizer.zero_grad()
    loss.backward()
    optimizer.step()
    
    # Step 6: every "z" iterations, pseudo-label unlabeled images if 
    # siam_vp's accuracy on the labeled set is 100%
    if iteration % z == 0 and siam_vp.train_accuracy == 100: 
        for j, image_unlabeled in enumerate(unlabeled_loader):
            # randomly select a labeled object and get $k$ reference viewpoints
            k = np.random.choice(len(labeled_set))
            reference_set = labeled_set.images[k]
            
            # get the viewpoint that aligns best with the unlabeled image
            viewpoint, confidence = siam_vp.predict(image_unlabeled, reference_set)
            
            # pseudo-label image j if the confidence and accuracy are high enough
            if confidence > 0.8: unlabeled_loader.dataset.label(j, viewpoint)
\end{minted}
\end{listing*}

\begin{figure*}[!h]
\begin{center}
   \includegraphics[width=1.0\linewidth]{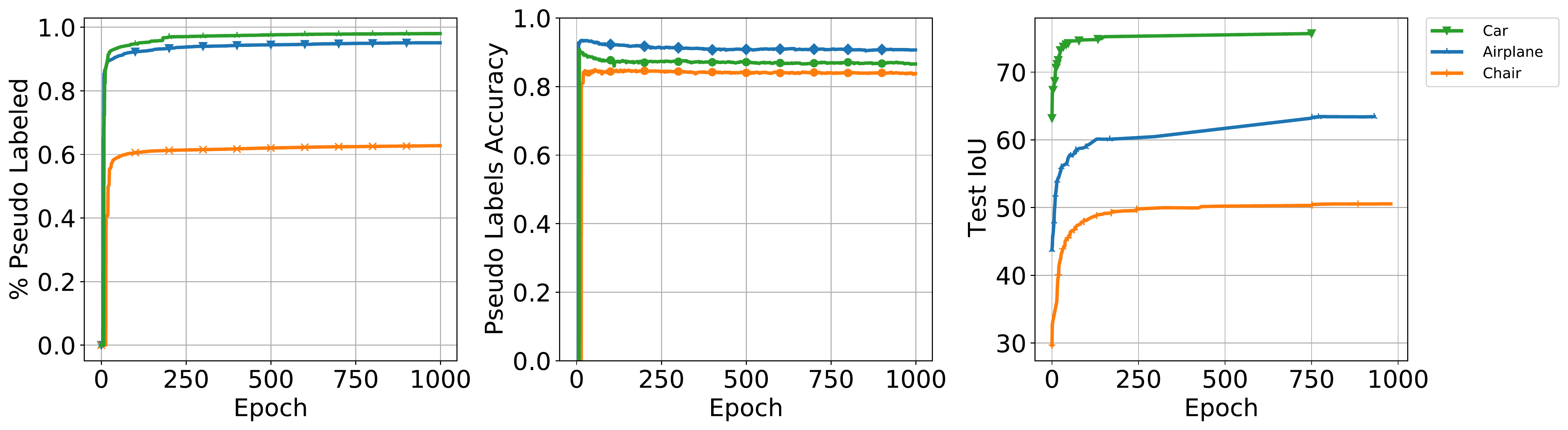}
\end{center}
   \caption{Results Across Pseudo-Labeling Cycles. The left plot depicts the percentage of unlabeled examples that has been pseudo-labeled by SiamVP. The center plot shows the accuracy of the pseudo labels across epochs. The right plot shows the 3D IoU on the ShapeNet's test set.}
\label{fig:pseudo}
\end{figure*}

\section{Experiments}
We evaluate the effectiveness of SSR for 3D reconstruction and viewpoint estimation. We first report the performance of SSR when trained with different numbers of examples and compare it with different baselines with different amounts of supervision. Then we study the evolution of SRR's performance across pseudo-labeling cycles and provide ablations to determine the influence of using  augmentations and metric learning. 

\subsection{Setup}
We experiment on ShapeNet v2 dataset~\cite{chang2015shapenet} which contains a rich collection of 3D computer-aided design (CAD) models of different classes of objects. Shapenet is widely used in recent research works related to 2D/3D data. We focus on 13 standard  categories of ShapeNet which are {\it Plane, Bench, Cabinet, Car, Chair, Display, Lamp, Speaker, Rifle, Sofa, Table, Phone, Vessel} as they offer enough objects to build a diversified image dataset and  they were previously used in a similar setup like~\citet{liu2019soft}. The number of objects in each class is shown in Table~\ref{tab:stats}. We benchmark with renderings provided by~\citet{kato2018neural}, where the 3D CAD object models were rendered at 24 uniform viewpoints to 64 × 64 images. We split the dataset into training, validation and testing sets accounting for 70, 10 and 20 percent of the whole dataset respectively, following~\citet{yan2016perspective}.  To simulate a semi-supervised setup, we further split the training set by randomly selecting a subset of the data to act as the labeled set, and the rest acting as unlabeled. We adjust the number of labeled samples in our experiments to show the effect of varying degrees of supervision. The splits are made on a object basis, that is, the different views from the same 3D model are either all labeled or all unlabeled.

We evaluate the 3D reconstruction accuracy by the 3D Intersection of Union (3D IoU) between the reconstructed images from every of the 24 viewpoint and ground truth 3D voxels of the test objects.

\subsection{Implementation Details}
We use the same structure and hyper-parameters as~\cite{liu2019soft} for mesh generation. Our network is optimized using Adam with $\alpha = 1 \cdot 10^{-4}$, $\beta_1 = 0.9$ and
$\beta_2$ = 0.999. We used a batch size of 64 and implemented our pipeline using PyTorch by building on the code found here.\footnote{\url{https://github.com/ShichenLiu/SoftRas}} For the SiamVP architecture, we used the Siamese network defined by~\citet{koch2015Siamese} with weights initialized using Kaiming Initialization~\cite{he2015delving}. We observed that this network attained better generalization than if we used the encoder architecture from~\citet{liu2019soft} which is more optimized towards constructing  3D meshes. We used a batch size of 32 for SiamVP and optimized the network using Adam with learning rate $10^{-4}$. These hyper-parameters were coarsely optimized using the validation set but better results would be likely achieved with more tuning on the validation set which would only strengthen the contributions made by this work. Note that we use the validation set for early stopping and we report the  test score achieved by the model that achieved the best result for the validation set in the 1000 epochs of training.

\subsection{Experimental Results}
\begin{figure*}[!h]
\begin{center}
   \includegraphics[width=0.87\linewidth]{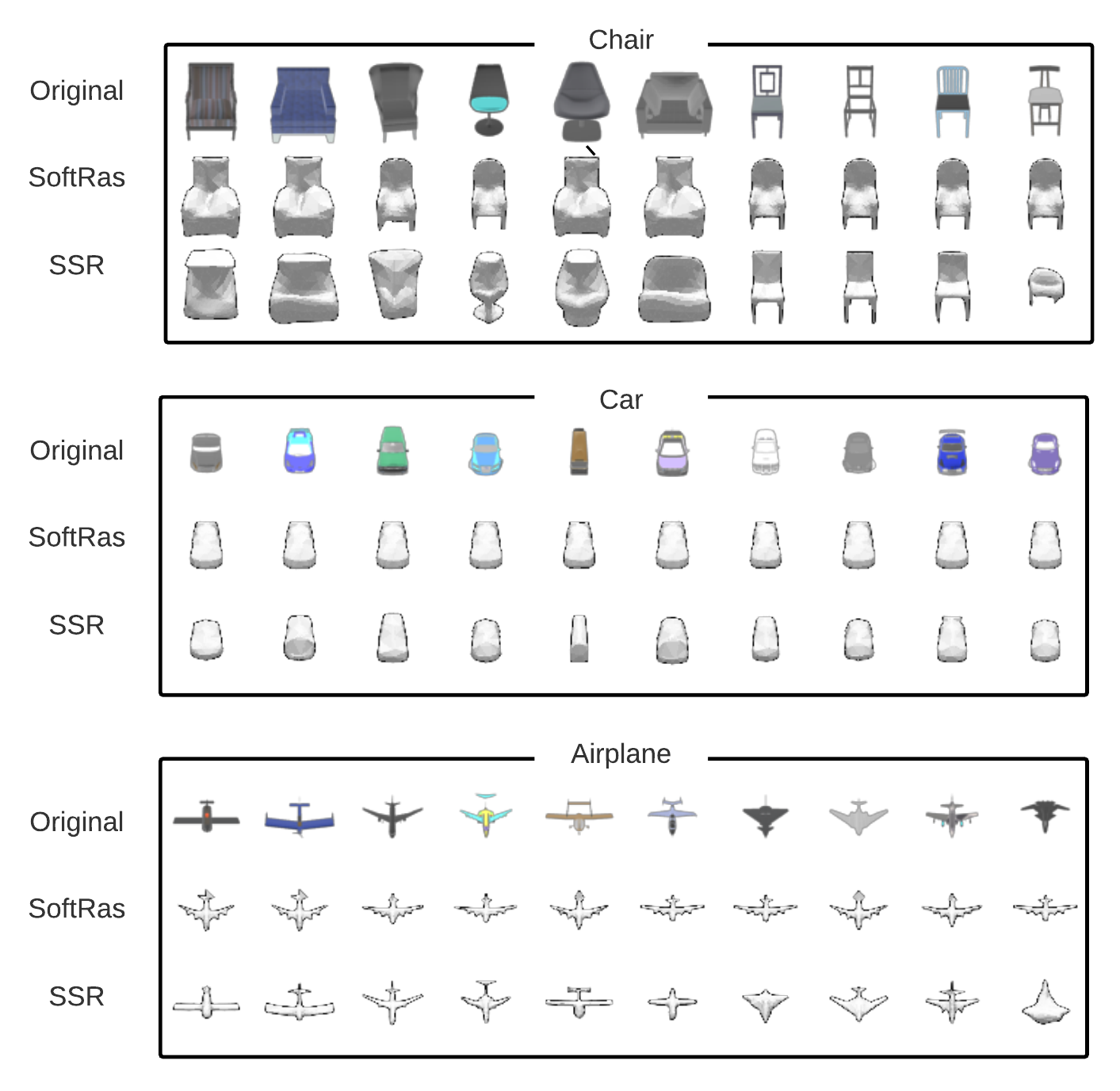}
\end{center}
   \caption{Qualitative results using only 2 labeled objects. SoftRas collapses to those two shapes mostly whereas, SSR: SiamVP and SoftRas are better at reconstructing the 2D image.}
\label{fig:qualitative}
\end{figure*}

\paragraph{Comparison with the baseline}
We compare the proposed SSR (SoftRas +SiamVP) against SoftRas for the cases where only few objects have been labeled. We report the results in Table~\ref{tab:results}. We observe that  by labeling only 2, 5, or 20 ShapeNet objects for each class with viewpoints, SSR consistently achieves a significantly better IoU test score than SoftRas. SSR uses the procedures described in Section~\ref{sec:methodology} which leverages the unlabeled set by pseudo labeling it with SiamVP. In Figure~\ref{fig:qualitative} we observe that SoftRas tends to overfit on the labeled set and fails to adapt to new objects. On the other hand,  SSR's reconstructions are more similar to their original counterpart. This result suggests that SiamVP can accurately predict the viewpoints of the unlabeled set which can be effectively used by SoftRas. 

\paragraph{Comparison with an unsupervised method}
MVC~\cite{tulsiani2018multi} is an unsupervised method that can learn camera parameters from objects given multiple views for each of them.  It does not use any viewpoint labels and it does not use differentiable rendering, thus it is a significantly different approach for 3D reconstruction than SoftRas. Its score is  reported in Table~\ref{tab:results}. Although MVC achieved good results considering that it used no viewpoint labels, SSR makes a significant improvement with only 2 objects being labeled, especially for the categories Plane, and Car. Unfortunately, MVC~\cite{tulsiani2018multi} does not report results for the remaining classes in order to make further comparison. The results however indicate that it is worth labeling extra objects and to use SSR to get a major increase in performance.

\paragraph{Comparison across number of labeled objects}
We compare the performance achieved from labeling 2, 5, or 20 objects from the training set in Table~\ref{tab:results}. We observe that SSR did not gain statistically significant improvement for classes like  \textit{Car, Vessel, and Rifle}. Our hypothesis is that many of the objects look similar in these classes, so labeling few extra objects might not provide rich information that SoftRas could use for better 3D reconstruction. On the other hand, for classes where objects are more diverse like \textit{Plane, Chair, Lamp, and Sofa}, SSR achieved big improvement with 5 labeled objects than with 2 labeled objects. We also see that there is a big boost in the \textit{mean} iou test score as the number of labeled objects increase, which suggests that SiamVP can still be effective with more training labels. 

\paragraph{Comparison with the upper bound}
In Table~\ref{tab:results} we also report results for SoftRas trained on the full labeled set. We observe that the performance gap between it and SSR (with 20 objects labeled) is small for most classes like \textit{Plane, Car, Display, Rifle}. Note that 20 objects is a small number compared to the full training set size which are shown in Table~\ref{tab:stats}. For the \textit{Plane} class, 20 only makes 0.7\% of the total number of Plane objects in the training set. This result indicates that for some classes few labels might be sufficient to get good results while for other classes, like \textit{Cabinet} and \textit{Speaker}, we would need more labels. This observation leads to an important future direction that could involve active learning to identify which objects are best to label for maximum information gain.

\paragraph{Comparison with a viewpoint classifier}
We also conducted an experiment  where we trained a viewpoint classifier using the same encoder as SiamVP that takes as input ShapeNet images and outputs the viewpoint of the image from a large list of hypothesis viewpoints. We observed a severe overfitting of the model on the labeled set and the results were worse than SoftRas that was trained only on the labeled set. Therefore, our metric learning approach with SiamVP is significantly more robust for classifying viewpoints than with a viewpoint classifier when the labeled set is small.

\paragraph{Results Across Pseudo-Labeling Cycles.} We investigate our model's performance as we grow the number of unlabeled examples that get incorporated to the training set with SiamVP's pseudo labels. Pseudo-labeling cycles occur every $2$ epochs, where every epoch consists of 200 training iterations. In Figure~\ref{fig:pseudo} we observe that the first set of labeled examples induce high accuracy gains on viewpoint prediction. However, as more viewpoints get labeled we observe some decrease in the viewpoint accuracy. For some shapes like \textit{Car}, the accuracy drops from 90\% to 88\% suggesting that SiamVP might start to overfit as the training progresses. 

On the other hand, we see that the accuracy plateaus above 80\% which indicates that SiamVP is robust to pseudo-labeling noise. Further, we see that for \textit{Plane} objects the ratio of pseudo-labeled samples is almost 100\% whereas for \textit{Chair} it is 60\%, suggesting that for some chair objects, SiamVP's predicted viewpoint probability was low. 

Note that SiamVP is training alongside SoftRas so at each pseudo-labeling cycle there is a chance that SiamVP labels a different set of unlabeled examples. We observed that this approach resulted in more accurate pseudo-labels across all object sets.

We also observe that the highest increase in the test IoU occurs in the first 20 epochs with a slight increase later in training. As mentioned in the implementation details, in the rest of the experiments we only report the test IoU at the end of their respective plot lines which correspond to the best validation model. 

\begin{table}[!t]
\caption{IoU results for various objects of ShapeNet.}
\label{tab:consis}
\footnotesize
\centering
\begin{tabular}{lllll}
\toprule
     Method &  Plane &  Bench & Cabinet & Car\\
\midrule
\midrule
SSR w/o augmentation &     0.45 & 0.30 & 0.43 & 0.64 \\
SSR (ours)&  0.55 & 0.44 & 0.48 & 0.75\\
\bottomrule
\end{tabular}
\end{table}

\paragraph{The importance of augmentation.}
Described in Section~\ref{sec:methodology}, the goal of augmentation is to help SiamVP generalize better as there is a limited number of viewpoints in the labeled set. In  Table~\ref{tab:consis} we compare the results with and without the augmentation for 4 ShapeNet classes where the number of labeled objects is 2. We observe that there is consistent improvement in the results when augmentation is added as it helps the model learn the 3D mesh while being invariant on the viewpoint of the 2D images. Note that for this approach we only used the rotation augmentation but other types of augmentation such as scaling, and translation can be used as well. Another augmentation technique to investigate is using a pretrained SoftRas to perform projections under novel viewpoints and use those projections as input to train a SiamVP.

\section{Conclusion}
We propose SSR, a semi-supervised soft rasterizer method that leverages unlabeled examples by using accurate pseudo labels. SSR consists of a viewpoint estimation component and a soft rasterization component. The viewpoint estimation is performed by a Siamese network called SiamVP that outputs whether two images have the same viewpoint. SiamVP slowly pseudo-labels images by carefully choosing only the most confident viewpoint predictions at each cycle. These viewpoint pseudo-labeling procedure leads to a signicant boost in the performance of SoftRas for reconstructing single 2D images to their 3D representation. Future directions for this work would include adopting this method for predicting viewpoints for objects in the wild where the environment is more challenging and less constrained than the ShapeNet dataset.

{\small
\bibliographystyle{abbrvnat}
\bibliography{egbib}
}

\end{document}